\documentclass[journal]{IEEEtran}
\usepackage[numbers,sort&compress]{natbib}
\usepackage{tcolorbox}
\usepackage{soul,xcolor}
\tcbuselibrary{breakable}
\usepackage{tabularx}
\usepackage{booktabs}
\usepackage{caption}
\usepackage{stfloats}
\usepackage{url,amsmath}
\usepackage{tabularray}
\usepackage{url}
\usepackage[hidelinks]{hyperref}
\usepackage[hidelinks]{hyperref}
\UseTblrLibrary{booktabs}
\usepackage[switch]{lineno}

\usepackage[inkscapelatex=false]{svg}
\setstcolor{red}
\tcbuselibrary{skins}
\newcolumntype{C}{>{\centering\arraybackslash}X} 
\def\tsc#1{\csdef{#1}{\textsc{\lowercase{#1}}\xspace}}
\tsc{WGM}
\tsc{QE}
\tsc{EP}
\tsc{PMS}
\tsc{BEC}
\tsc{DE}

\begin{document}

\title{\begin{huge} \textbf{RAISE}: A \textbf{\underline{\smash{R}}}obot-\textbf{\underline{\smash{A}}}ssisted Select\textbf{\underline{\smash{I}}}ve Disassembly and\\ \textbf{\underline{\smash{S}}}orting System for \textbf{\underline{\smash{E}}}nd-of-Life Phones
\end{huge}}

\author{Chang~Liu$^{1}$,
        Badrinath~Balasubramaniam$^{2}$,
        Neal~A. Yancey$^{3}$,
        Michael~H.~Severson$^{3}$,
        Adam~Shine$^{4}$,
        Philip~Bove$^{4}$,
        Beiwen~Li$^{2}$,
        Xiao~Liang$^{5,*}$,
        and~Minghui~Zheng$^{1,*}$
\thanks{This material is based upon work supported by the REMADE Institute, USA (21-01-RM-5083). Any opinions, findings, and conclusions or recommendations expressed in this material are those of the authors and do not necessarily reflect the views of the sponsor.}
\thanks{$^{1}$Chang Liu and Minghui Zheng are with the J. Mike Walker '66 Department of Mechanical Engineering, 
Texas A\&M University, College Station, TX 77843, USA. 
Emails: \href{mailto:changliu.chris@tamu.edu}{changliu.chris}, 
\href{mailto:mhzheng@tamu.edu}{mhzheng@tamu.edu}.}
\thanks{$^{2}$Badrinath Balasubramaniam and Beiwen Li are with the College of Engineering, University of Georgia, Athens, GA 30602, USA. 
Emails: \href{mailto:bb2@uga.edu}{bb2}, 
\href{mailto:beiwen.li@uga.edu}{beiwen.li@uga.edu}.}
\thanks{$^{3}$Neal A. Yancey and Michael H. Severson are with the Idaho National Laboratory, Idaho Falls, ID 83415, USA. 
Emails: \href{mailto:neal.yancey@inl.gov}{neal.yancey}, 
\href{mailto:michael.severson@inl.gov}{michael.severson@inl.gov}.}
\thanks{$^{4}$Adam Shine and Philip Bove are with the Sunnking Sustainable Solutions, Brockport, NY 14420, USA. 
Emails: \href{mailto:ashine@sunnking.com}{ashine}, 
\href{mailto:pbove@sunnking.com}{pbove@sunnking.com}.}
\thanks{$^{5}$Xiao Liang is with the Zachry Department of Civil and Environmental Engineering, 
Texas A\&M University, College Station, TX 77843, USA. 
Email: \href{mailto:xliang@tamu.edu}{xliang@tamu.edu}.}
\thanks{$^{*}$Corresponding Authors.}}

\maketitle
\begin{abstract}
End-of-Life (EoL) phones significantly exacerbate global e-waste challenges due to their high production volumes and short lifecycles. Disassembly is among the most critical processes in EoL phone recycling. However, it relies heavily on human labor due to product variability. Consequently, the manual process is both labor-intensive and time-consuming. In this paper, we propose a low-cost, easily deployable automated and selective disassembly and sorting system for EoL phones, consisting of three subsystems: an adaptive cutting system, a vision-based robotic sorting system, and a battery removal system. The system can process over 120 phones per hour with an average disassembly success rate of 98.9\%, efficiently delivering selected high-value components to downstream processing. It provides a reliable and scalable automated solution to the pressing challenge of EoL phone disassembly. Additionally, the automated system can enhance disassembly economics, converting a previously unprofitable process into one that yields a net profit per unit weight of EoL phones.
\end{abstract}

\IEEEpeerreviewmaketitle

\section{Introduction}
E-waste presents a global challenge due to its rapid growth, high resource value, and the severe environmental and health risks from improper recycling and hazardous substances \cite{liu2023global, kar2025review, xavier2023comprehensive}. Global e-waste surged to a record 62 million tonnes in 2022 and is expected to reach 82 million tonnes by 2030 \cite{Globalewaste2024}. Recycling converts e-waste components into valuable raw materials, which is critical for addressing the escalating e-waste problem and supporting a sustainable circular economy \cite{vulsteke2024meaning, lu2023state,kumari2022critical, ueberschaar2017assessment, tutton2022pre, mccann2015solving}. Nevertheless, only 22.3 \% of e-waste was recorded as recycled in 2022 \cite{Globalewaste2024}. The high human labor cost and health risk concerns are the major challenges associated with the recycling process \cite{ceballos2016formal}. A recent study reported that labor contributes over 90 \% of e-waste recycling costs in Australia \cite{dias2019ensuring}.

Recycling End-of-Life (EoL) electronic products, such as phones, laptops, desktops, etc., is increasingly important because of their shortened lifecycles and rapid technological advancements \cite{yao2018integrated, zhang2024selection, vidal2023towards, maani2023estimating}. EoL phones stand out due to their short lifecycle of 1.98 years \cite{prabhu2023disposal} and rapidly growing production volumes, which amplify the urgent need for efficient recycling solutions \cite{brundl2024towards}.

Disassembly is a vital and fundamental step in recycling, which enables efficient separation of valuable materials, enhances recovery rates, and ensures safe handling of hazardous components \cite{habib2019product, golev2019estimating, oke2024discarded}. Nevertheless, modern electronic devices are primarily designed for manufacturing efficiency and user convenience, such as waterproof and lightweight designs, rather than for ease of EoL disassembly \cite{abuzied2020review}. These intricate and compact designs, particularly for EoL phones, significantly complicate the disassembly process. Additionally, the quality and type variations of collected EoL phones introduce additional uncertainties, which require high-level decision-making ability \cite{lu2023state}. Manual disassembly remains the predominant method in practice because of humans' adaptability and intelligence. However, it is labor-intensive, inefficient, and exposes workers to potential hazards and unsafe working environments \cite{duflou2008efficiency, movilla2016method, rautela2021waste}.

These constraints highlight the urgent need for advanced automated disassembly systems tailored to EoL phones. The scalability, sustainability, and efficiency of the system are essential to address the increasing volume of EoL phones and their growing design complexity. Although some automated disassembly systems for vehicles \cite{tolio2017design}, batteries \cite{hellmuth2021assessment}, televisions \cite{bogue2019robots}, panels \cite{ALR}, etc., have been investigated, research specific to EoL phones remains limited. Apple's ``Daisy'' system \cite{Appledaisy} is only designed for iPhone, Takao's high-speed system \cite{ueda2024automatic} only focuses on EoL phone battery separation without sorting, and Figueiredo's \cite{figueiredo2018high} robotic system is particularly subject to the prying operation during disassembly. These systems are highly customized and have high capital costs, which hinder their potential deployment. Collectively, these factors result in a gap between industrial needs and technological advancements.

\begin{figure*}[!t]
    \refstepcounter{figure}
    \addcontentsline{lof}{figure}{\protect\numberline{\thefigure}sample}
    
        \begin{minipage}{\linewidth}
            \begin{center}
                        \includegraphics[width=0.95\linewidth]{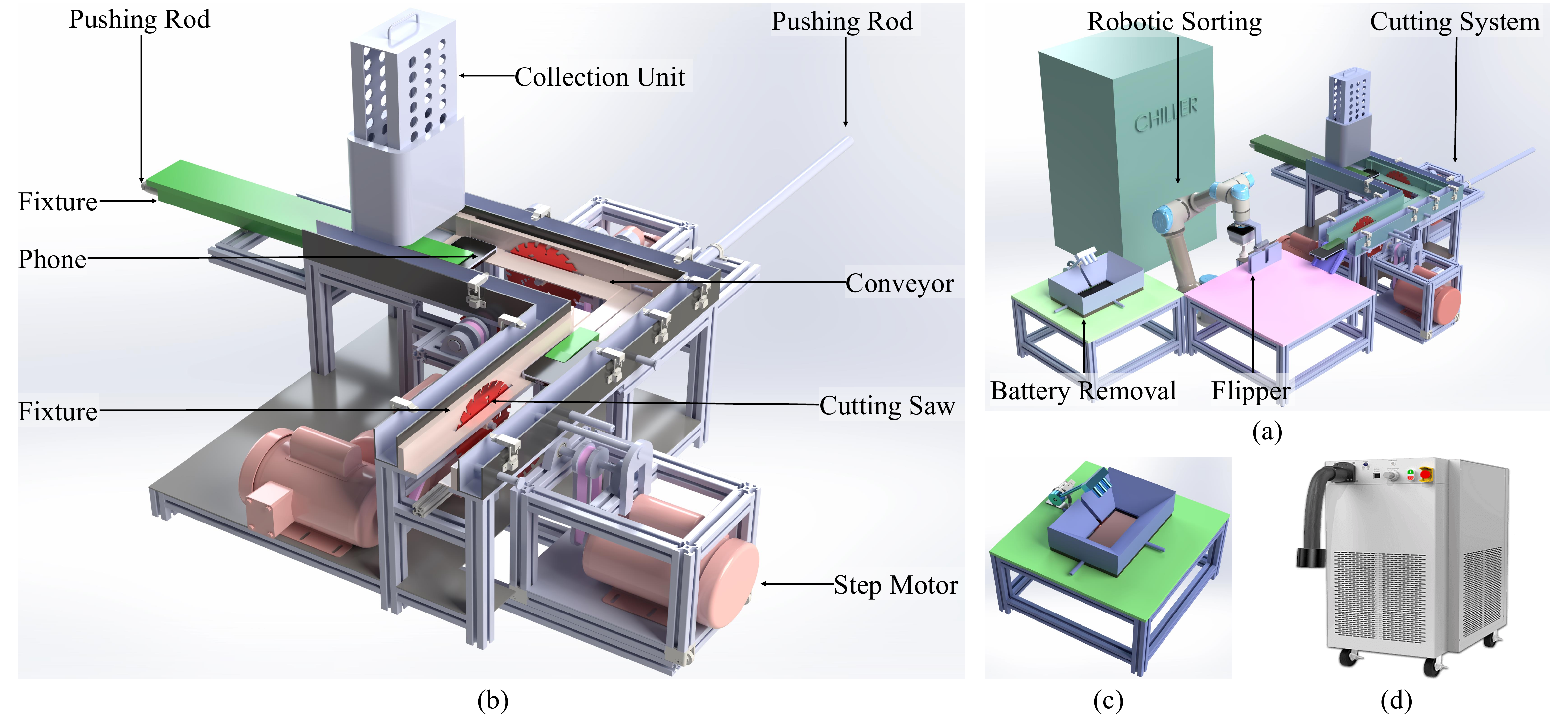}
            \end{center}
            \vspace{4pt}
            \noindent\textbf{Figure \thefigure:} (a) Overview of the proposed semi-deconstructive automated selective disassembly and sorting system for EoL phones. It consists of an adaptive cutting system, a robotic sorting system, and a battery removal system. (b) The adaptive two-way cutting system is equipped with four sets of step motor-driven cutting saws, which are used for rapid and secure cutting. The phones are initially stored in the collection unit. Then, two pushing rods work with the adaptive fixture to transport the phone along the L-shaped conveyor for cutting securely. (c) The swinging hammer mechanism is used to separate the battery from the phone. (d) The chiller system is used to freeze the phone part with the battery.
        \end{minipage}
    \label{systmemodule}
\end{figure*}

To address the gap and meet practical needs, this paper presents a robot-assisted selective disassembly and sorting system for EoL phones, as illustrated in~\ref{systmemodule} (a). It is a fully automated system without human intervention, in which the robot works with the adaptive cutting machine to automate the entire disassembly and sorting process.
This system focuses on the component functional layer level disassembly rather than the individual internal parts, like RAM chips and vibrators. This target is a defined techno-economic endpoint, which is designed to maximize profit. The system can separate and safely process the low-value and hazardous components, including the plastic cover, screen, film, and battery. Simultaneously, it sorts the high-value component layers for downstream detailed recovery. This strategy finds a balance between efficiency and cost-effectiveness, which can further be compatible with the existing workflows. To align with the strategy of the EoL phone disassembly, this system features several key contributions that distinguish it from existing research and industry practices.

\begin{itemize}
    \item \textbf{Automated Cutting System}: An adaptive cutting system that can process diverse phone models for effective and risk-free layer separation with minimum damage to the main components, which can make the phone ready for the subsequent processes.

    \item \textbf{Vision-Based Robotic Sorting System}: A hand-eye vision and deep learning based robotic sorting system, which achieves 98.9\% accuracy in real-time decision making and disassembly process planning. This system determines the consistency and quality of outputs.

    \item \textbf{Battery Removal System}: A thermal air chilling system with a mechanical swinging hammer to separate batteries quickly and safely.

    \item \textbf{Dataset}: A dataset including over 2300 images with pixel-level segmentation masks from five different types of phones. The dataset can serve as a baseline for segmentation model evaluation in vision-guided EoL product disassembly systems.
\end{itemize}

Additionally, the whole system doesn't require prior knowledge or costly equipment to handle the uncertainty in the disassembly process. Overall, the automated system provides an efficient, scalable, and adaptive solution for selective layer-level EoL phone disassembly and sorting.
This work improves current recycling practices by demonstrating the viability of an automated system to meet the increasing demands for sustainable EoL product management.

\section{Related Work}
This section reviews existing disassembly approaches for different EoL products and emphasizes the need for automated disassembly solutions for EoL phones.
Complete disassembly can fully dismantle all components, but is usually costly and unnecessary for most EoL products \cite{ramirez2020economic}. Selective disassembly presents a practical compromise by targeting the high-value and hazardous components, then effectively sorts them for the downstream process \cite{smith2016partial, gao2023data, li2013selective, kiyokawa2022challenges}. Zhang et al.~\cite{zhang2024selection} demonstrate that selective disassembly can increase processing efficiency by 48\% in EoL product recycling. Non-destructive disassembly retrieves components without damaging them, but it relies heavily on humans and is often impractical \cite{hjorth2022human, guo2025exploring}. Semi-deconstructive disassembly permits minor damage to fasteners and low-value parts, which enables efficient recovery of valuable components \cite{lee2024review}. Given these advantages, combining selective and semi-destructive disassembly stands out as a highly effective approach for automated systems \cite{edis2022mixed}.

Manual disassembly can handle the uncertainty of EoL products with intelligent decision-making and manipulation abilities, which is challenging for automated systems \cite{kopacek2006intelligent, carrell2009review}. However, it remains time-consuming and economically inefficient. As EoL product volumes escalate, the need for semi- or fully automated disassembly systems is gaining more attention \cite{shahjalal2022review, movilla2016method, ALR}.

Human-robot collaborative disassembly (HRCD) provides a dream combination of humans' decision-making abilities for uncertainties and robots' repetitive reliability \cite{liu2019human,dhanda2025reviewing}. Effective task allocation and disassembly sequence planning (DSP) are essential to EoL products \cite{lee2022robot,zafar2024exploring}. DSP has been proven to enhance safe disposal \cite{huang2020case}, maximize recyclability \cite{he2024disassembly}, improve efficiency \cite{hu2024ontology, huang2021experimental}, and support informed decision-making in remanufacturing \cite{liu2019human}.

HRCD has attracted wide attention in EV battery disassembly because of its high recovery value and standardized structure design, which reduces uncertainty \cite{chu2023human}. Manufacturers usually dominate this by leveraging detailed design data to optimize robotic task planning \cite{hellmuth2021assessment}. Li et al. review how artificial intelligence can empower the HRCD in EV battery \cite{li2024end}. Wegener et al. \cite{wegener2015robot} propose a robotic workstation to assist humans in simple and repetitive tasks for EV battery screw removal. Gerbers et al. \cite{gerbers2018safe} implement HRCD in lithium-ion battery removal. Bdiwi et al. \cite{bdiwi2017disassembly} present a concept of using HRCD in vehicle motor disassembly.

Recent HRCD research is focusing on enabling robots to make real-time decisions in uncertain environments. Lee et al. proposed a HRCD task allocation and DSP algorithm to reduce the hard disk drive disassembly time while ensuring safety constraints \cite{lee2022task}. Tian et al. \cite{tian2023optimization} propose an optimization-based human behavior model and expand their work by using the transformer-based diffusion model to realize a real-time human motion prediction in the HRCD environment \cite{tian2024transfusion, zhang2024early}. Saran et al. propose a DSP algorithm on the EoL fuel pump \cite{parsa2021human}. Zhout et al. \cite{zhou2022stackelberg} develop a new stackelberg model to address the uncertainty in screws in HRCD. Liu et al. \cite{liu2024hybrid} present a real-time motion planning algorithm in HRCD tasks based on an online manipulator reconfiguration dataset. However, the HRCD practical implementation is still challenging due to three critical criteria: efficiency, cost, and adaptability \cite{lee2024review}. They are especially vital for EoL products with low recycling or remanufacturing value, such as phones, where labor dominates the cost \cite{zhang2019remanufacturing, lu2023state}. Jiang et al. \cite{jiang2020data} present a cost prediction model to analyze the cost of remanufacturing different EoL products. EoL phones receive minimal attention in HRCD due to their low recycling value, inefficiency, and high labor cost. Additionally, the large volume of EoL phones underscores the need for an automated disassembly system with minimal human intervention.

Automated disassembly systems offer a sustainable approach by integrating robotics and computer vision to perform disassembly tasks with high efficiency and precision \cite{ramirez2020economic, palmieri2018automated}. However, both research and practice are facing challenges from product uncertainties and system capabilities such as perception, tooling, operation, reliability, and cost \cite{chen2021automated, foo2022challenges}. The uncertainties include variations in the design, condition, structure, and positioning. Perception typically relies on sensors and cameras to handle the variety and condition of the product \cite{bonci2021human}. Tooling, operation, and reliability determine whether the system can execute tasks with a consistently high success rate based on the perception information, and then sort the components accurately and efficiently. Overall, a successful automated disassembly system must integrate perception and tools, handle uncertainties, make informed decisions, and deliver the target components. Additionally, low deployment costs are essential to ensure scalability and economic viability for local facilities.

The robotic vision system has been widely explored with deep learning methods to enhance its perceptual capabilities. Sarswat et al. \cite{sarswat2024real} use convolutional neural networks (CNN) for real-time e-waste detection and achieve 94\% accuracy. Zhang et al. \cite{zhang2023automatic} present a tool recommendation system based on the YOLOv4 architecture, which can recommend the appropriate tool for agents. Sterkens et al. \cite{sterkens2021detection} utilize a deep learning method to detect lithium-ion batteries from X-ray images of e-waste. Similarly, Ueda et al. \cite{ueda2024line} employ X-ray transmission scanning to capture images of the inner structure of EoL products, and then utilize a deep learning model to detect the presence of batteries and screws inside. Jahanian et al. \cite{Jahanian_2019_CVPR_Workshops} propose a deep neural network model for multi-task instance segmentation purposes, which can detect the boundaries of small electric components on phone PCB boards. However, deep learning models are facing challenges due to dataset limitations, including quality, annotation, and representational biases, which directly impact their performance and applicability to use them in generalized practice \cite{shwartz2022tabular,paullada2021data, dong2025benchmarking}.

Most existing datasets are used to separate e-waste devices from solid waste streams \cite{lin2014microsoft, OpenImages, yang2016classification, lu2022computer, ekundayo2022device, iliev2024proposal}. There is currently no standardized, component-level instance segmentation dataset for EoL disassembly purposes. This absence significantly limits the development and fair evaluation of deep learning models in this field. Jahanian et al. \cite{Jahanian_2019_CVPR_Workshops} introduced a dataset with 553 images for small phone components. However, it was only used in their own model evaluation and not adopted more broadly. Similar limitations are seen in datasets for hard disks \cite{rojas2022deep}, laptops \cite{bassiouny2021comparison}, and screws \cite{zhang2023automatic, kalitsios2022vision}.

Although the HRCD in EoL products is developing rapidly, there is a growing shift toward fully automated disassembly and sorting systems. Several industrial implementations have demonstrated the viability of such systems in structured, large-scale environments. Renault and Indra operate disassembly sites capable of processing approximately 25 ELVs per day, achieving a 95\% recycling rate of total vehicle mass \cite{tolio2017design}. In consumer electronics recycling, VEOLIA's RoboTele system can dismantle up to 500,000 flat-screen televisions per year. Similarly, Votechnik's ALR4000 system can process over 60 flat panels per hour \cite{bogue2019robots, ALR}.

\begin{figure*}[htbp]
	\centering
	\includegraphics[width=0.95\textwidth]{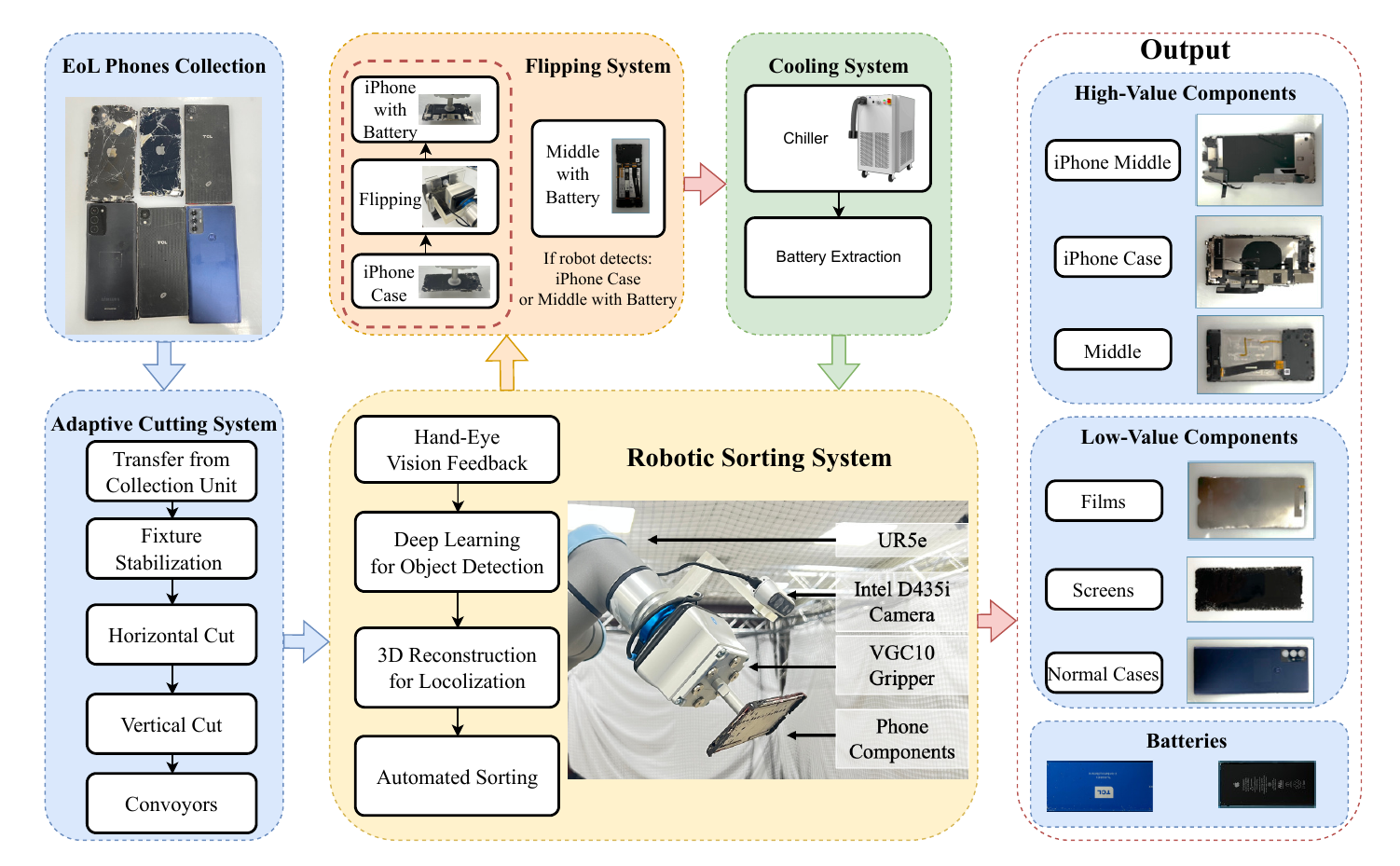}
	\caption{The whole automated disassembly process by our proposed system. The EoL phone is designed to go through the cutting process, the sorting system, and the battery removal system. All the valuable components of the phone are delivered to the next disassembly stage.}
	\label{wholesystem}
\end{figure*}

As to automated disassembly systems for EoL phones, several additional complexities exist, including larger product variability, increased uncertainty, lower recycling value, and significantly higher processing volumes. 

\begin{itemize}
    \item Apple's ``Daisy'' system exemplifies brand-specific automation in this area, which can fully disassemble up to 200 iPhones per hour \cite{Appledaisy}. However, it depends heavily on proprietary design-stage data, making it inaccessible to external research or facilities. Moreover, it is limited to undamaged iPhone models and does not support other brands.
    \item Takao's high-speed selective disassembly system integrates X-ray scanning and deep learning methods for battery and screw detection \cite{ueda2024automatic}. It can process up to 600 smartphones per hour, achieving an 88.3\% success rate, by utilizing freezing and semi-destructive cutting methods. However, the system focuses only on battery removal and leaves other components clustered together, which complicates downstream disassembly.
    \item Figueiredo's robotic system focuses on the non-destructive automated disassembly for EoL phones \cite{figueiredo2018high}. This system integrates a specially designed gripper for the prying operation and introduces a high-level decision-making algorithm to deal with the uncertainty in real-time. However, this system is limited to the prying operation and doesn't address the requirements of the current EoL phone disassembly practices. 
\end{itemize}

Despite recent advances, most existing systems face significant barriers to scalability. High capital costs and system complexity prevent their practical use in local recycling centers. These centers often operate with limited space, labor, and resources. As a result, automated disassembly systems are rarely deployed in decentralized settings. They rely on the global centralized collection and transportation of intact EoL phones. This increases cost and reduces overall economic feasibility.

\section{Proposed Disassembly and Sorting System}

The proposed disassembly system balances the deployment cost, regulatory compliance, and operational flexibility, which aligns with the needs of current EoL phone recycling practices. The entire disassembly system comprises three subsystems that construct a scalable and fully automated solution together. The semi-destructive adaptive cutting system enables the safe and efficient separation of phone layers across various models. The robotic sorting system applies vision and deep learning methods to perform real-time decision-making and planning for sorting. The sorting system separates high-value components for downstream recovery and directs low-value parts to recycling, significantly reducing labor costs through robotic automation. It also works in conjunction with the battery removal system, which utilizes thermal chilling and mechanical separation to satisfy the environmental and safety regulations for lithium-ion battery processing. The detailed description of these three subsystems is provided below.

\subsection{Adaptive Cutting System}
The adaptive cutting system is developed to remove the connections and fasteners between phone layers, regardless of the type and size of different phones. Four mechanical methods are tested and evaluated based on three criteria: safety, efficiency, and output quality. While water jet cutting provides a clean, safe, and precise cut, an 18-minute processing time per phone is considered not practical. Hydraulic press cutting reduces the process time to 2 minutes. However, the batteries are often found damaged after the process, which raises safety concerns in practice. The band saw machine can accomplish the cutting on four sides within 40 seconds. However, due to the spatial constraints of the band saw, the process requires the phone to be transferred sequentially three times for its four-sided adaptive cutting. The repeated placements introduce additional precision requirements, and the accumulated error can compromise the output quality for subsequent robotic sorting. In contrast, the two circular saws can be arranged in parallel as a pair for two-sided simultaneous cutting. With the integration of passive and adaptive fixtures, phones with different sizes only need to be transferred once between two circular saw pairs to achieve four-sided cutting during the process. This method is selected because it achieves safe and consistent high-quality clean cutting output. It also demonstrates the highest efficiency within 30 seconds for different EoL phones across all tested methods. The combination of reliability, efficiency, and adaptivity makes the selected circular saw method align with the requirements of our cutting system. The whole system is shown in Figure~\ref{systmemodule} (b).

\subsection{Automated Robotic Sorting System}
EoL phones are separated into different component layers after the cutting system. These layers need to be sorted based on their value for subsequent disassembly steps. Manual handling of these raises both efficiency and safety concerns. Fragile parts, such as screens or iPhone cases, may crack during cutting and leave sharp debris. It poses a risk of injury to unprotected workers. Wearing protective gloves can prevent injury, but makes it difficult to grip thin layers. These challenges underscore the need for a robotic system that is capable of safely and efficiently detecting and sorting components. To address this issue, we built an automated robotic system to sort component layers at this stage. We use Universal Robot (UR5e) as the manipulator and Onrobot VGC10 vacuum gripper as the end-effector. The vacuum gripper can efficiently handle the thin layers of the components after cutting. It can directly attach the components from the top and use the suction force to lift them. Moreover, we use the Intel RealSense D435i camera with a customized mount as a hand-eye system for real-time object detection. With all the features above, this system can sort the component layers of EoL phones in fully automated mode.

\subsubsection{Hand-Eye Vision System}
In this study, a standard hand-eye calibration process is performed to obtain the spatial relationships between the camera vision and robot execution. The camera's intrinsic and key transformation parameters are calculated during this process, which is used to localize the component for the robot from 2D images.

\subsubsection{Deep Learning for Components Detection}
\begin{figure}[htbp!]
	\centering	\includegraphics[width=.45\textwidth]{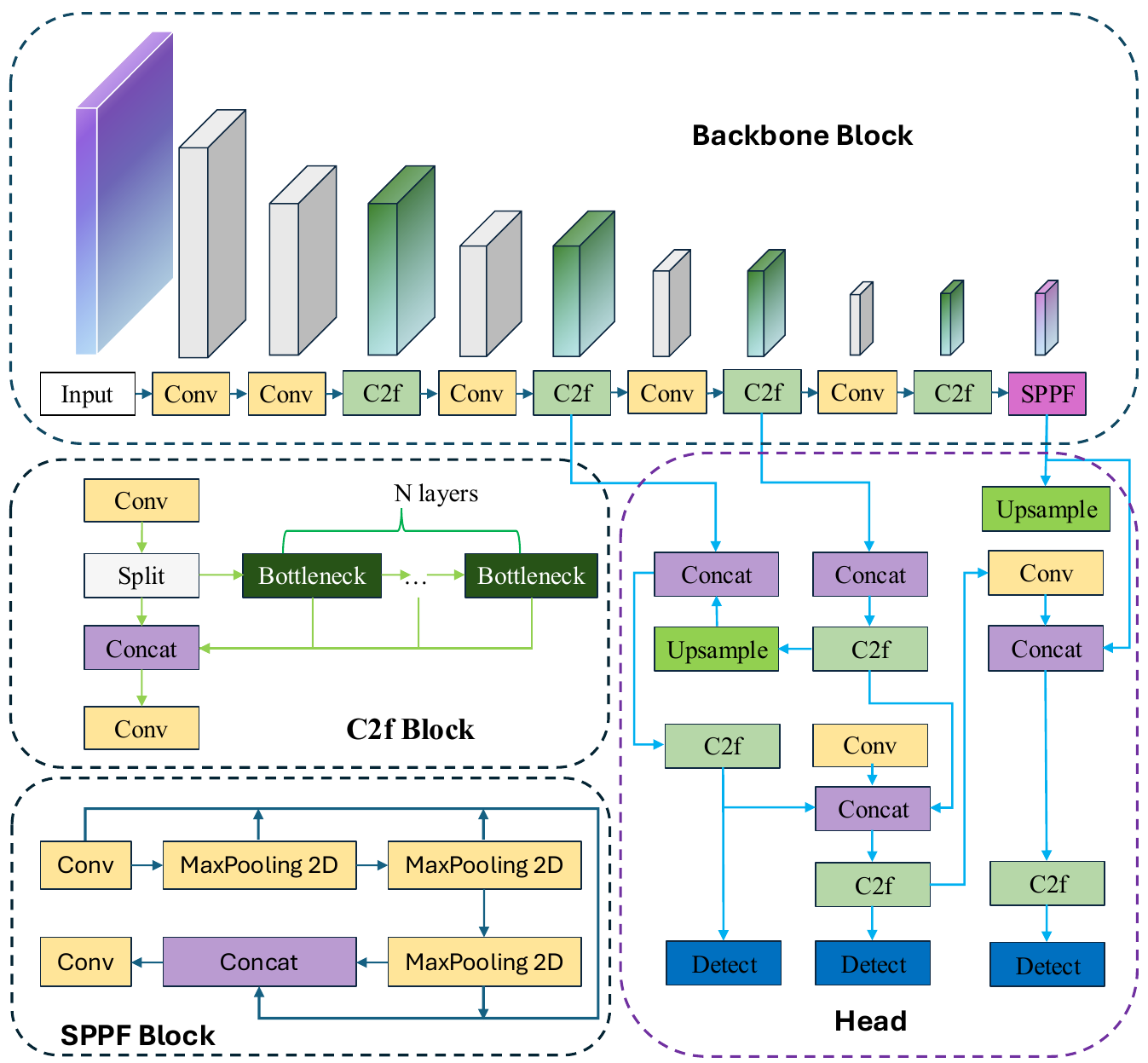}
        \caption{The architecture of YOLOv8 segmentation and detection neural network.}
        \label{yoloarc}
\end{figure}

Object detection and segmentation are crucial for accurately identifying the various layers of EoL phones after they have been cut. YOLOv8 framework is selected in this work due to its capability to perform single-stage object detection and instance segmentation with both high accuracy and real-time efficiency \cite{varghese2024yolov8}. The structure of YOLOv8 can be found in Figure~\ref{yoloarc}.
To adapt YOLOv8 for this specific disassembly task, transfer learning is employed using pre-trained weights and fine-tuned with our custom dataset of different EoL phone component layers. This approach can make YOLOv8 well-suited for our automated sorting workflows to differentiate parts accurately.

\subsubsection{Sorting Process}
\begin{figure*}
	\centering
	\includegraphics[width=.95\textwidth]{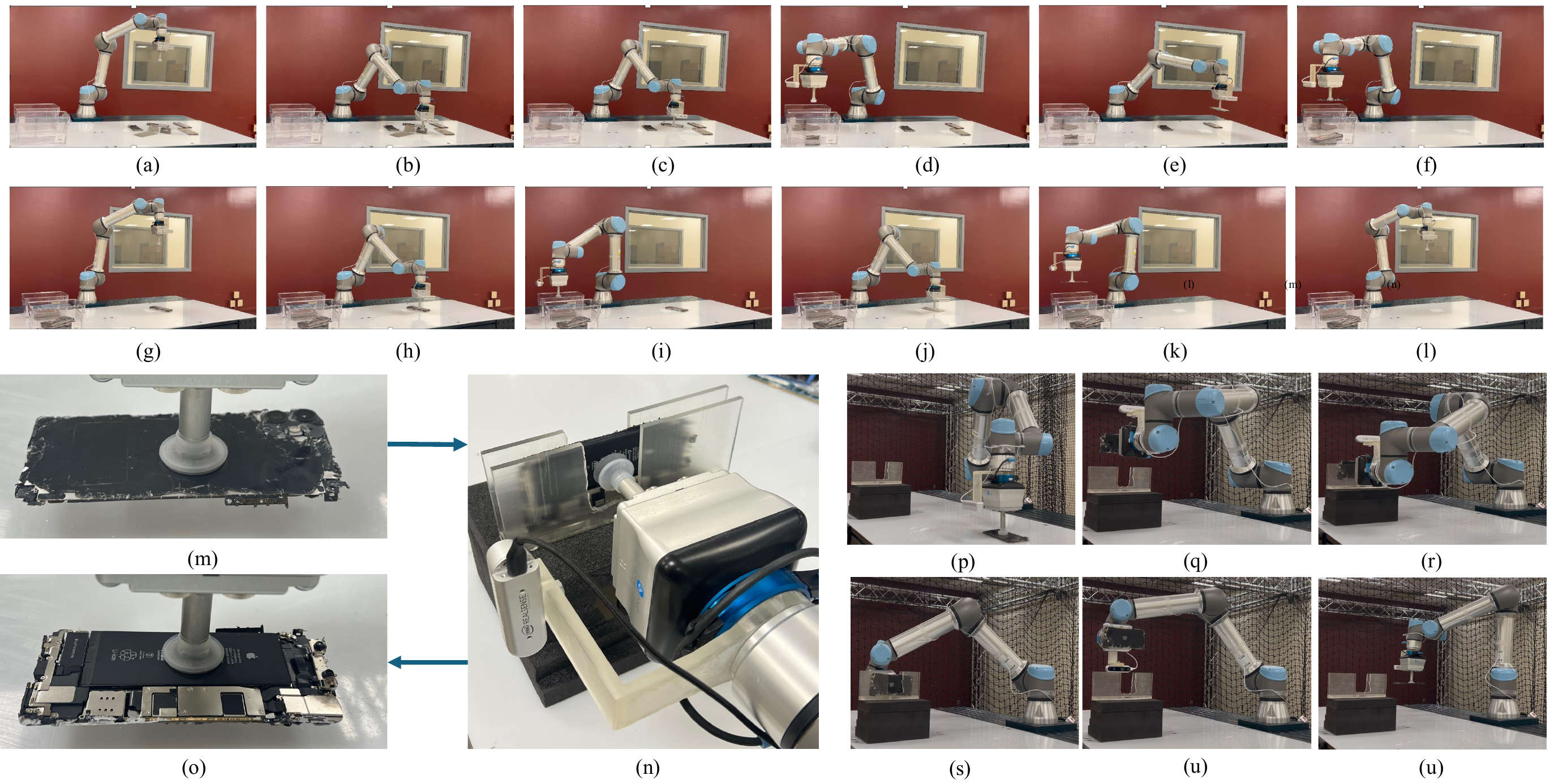}
        \caption{
        The demonstration of the real-time sorting process of the vision-based robotic system. In the first scenario, it sorts all the components on the table according to their values as (a-f). In the second scenario, it sorts the stacked components delivered by the cutting system as (g-i). Regarding the flipping process, the robot first picks up the iPhone case as (m), then inserts it into the case holder vertically as (n), and subsequently retrieves the iPhone case from another direction as (o). We also demonstrate the real-time flipping process as (p-u).}
        \label{flipping}
\end{figure*}

After cutting, all components are placed face down and stacked. The vision system detects all components on the table, along with their orientation.
Components fall under standard categories such as case, film, and screen, which are sorted directly into a designated low-value bin. The two experimental sorting processes are illustrated in Figure~\ref{flipping} (a-i). When the system detects the iPhone case, it initiates the flipping process, as shown in Figure~\ref{flipping} (m-u). The process redirects the battery inside the iPhone case to face up, which makes it suitable for the battery extraction process. The flipped iPhone case and the middle layer are then transferred to the cooling system for battery extraction. Once the separation process is completed, the robot retrieves the remaining parts and sorts them into their respective bins.

\subsection{Battery Removal Cooling System}
Phone batteries are typically secured inside the main frame with strong glue. Various chemical and mechanical methods are tested to evaluate their effectiveness in detaching the batteries. Initially, only mechanical methods are applied. However, this approach could damage the batteries easily and potentially cause a fire. Then, the freezing techniques are tested, which can significantly reduce fire risk and effectively break the glue connection. The most effective method is to utilize a chiller to freeze the adhesive connection. Then we use a mechanical system to extract the battery from the frame.
 
The MPI ThermalAir TC-100 is selected to provide the required cooling capability, as shown in Figure~\ref{systmemodule} (d). It is a high-capacity thermal air chilling system that can deliver $-80^\circ\text{C}$ air at the output at up to 24 standard cubic feet per minute (SCFM). The system provides a direct stream of continuous, clean, dry, and cold air to the battery. The glue weakens significantly when frozen below $-17.78^\circ\text{C}$. A mechanical swinging hammer is then used to strike the frame, which can easily break the adhesive and extract the battery from the main frame, as shown in Figure~\ref{systmemodule} (c).

\section{Experiments and Results}
\subsection{Cutting System}
The system automatically transfers collected EoL phones from the collection unit to the cutting system. The phones are placed inside the collection unit with the correct orientation. Initially, the phone is stabilized on the first section of the L-shaped conveyor using the adaptive fixture. The fixture's spring mechanism can firmly secure phones of different sizes during cutting. Notably, the circular saw blade module could move with the fixture to match the size for simultaneous cutting. The phone moves with the pushing rod to finish the first cutting process, then the phone is released and controlled by the second pushing rod and fixture. The fixture controls the cutting depth and maintains $2mm$ offset from the edge on each side.

\subsection{Automated Sorting System}
\subsubsection{Dataset}{\label{Dataset}}
The dataset is constructed for phone component layer detection tasks. We select five different types of phones across two different design categories: iPhone and Android-based phones. The iPhone SE and iPhone 11 represent iPhone design variants in different sizes. The Samsung Galaxy A03S, Motorola G Pure, and TCL 30E are selected to represent the general Android-based phone with a battery-centric modular structure. They come from three major manufacturers in the world with notable market share, excluding Apple. Together, these five types of phones contain the majority of the structural designs on the market. We select 120 phones with different physical conditions from these five types. We further process all the phones with the cutting system.

The layer components obtained after cutting have similarity across different models. They are categorized into five value-based classes: normal case, middle layer, screen, film, and iPhone case. The Intel RealSense D435i camera is used to capture photos from different angles, heights, and light conditions to simulate potential working situations. We collect 2376 images with a 1920x1080 pixels resolution for different phone layer components. We use an online labeling platform RoboFlow \cite{Roboflow}, which is designed for YOLOv8 model data preparation. Compared to the previous datasets for object detection purposes, we label all the layer component features for instance segmentation purposes.

\begin{figure*}
	\centering
	\includegraphics[width=.92\textwidth]{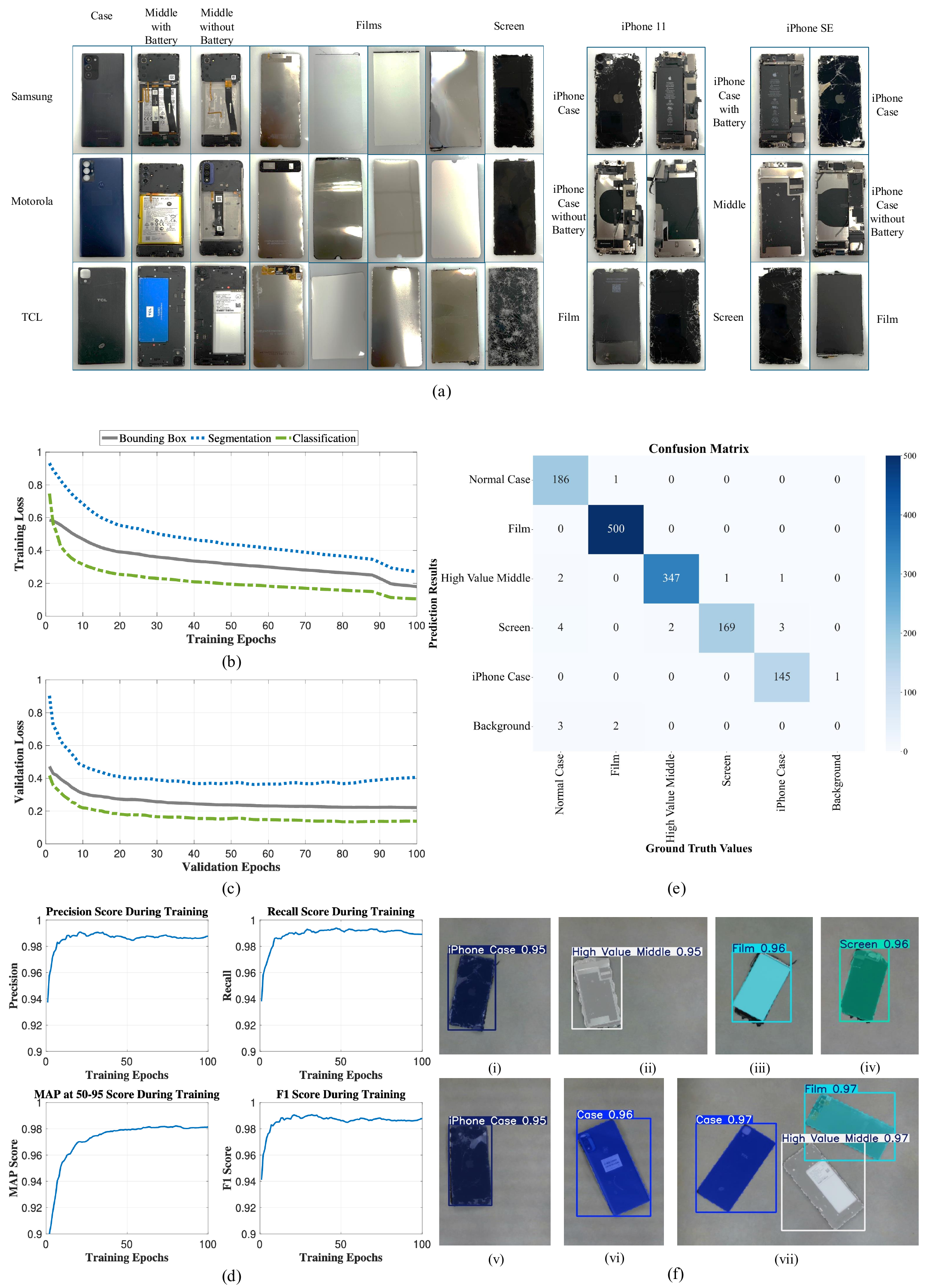}
	\caption{(a) The component layers of each brand of phones after cutting. (b) The training loss changes during the training process. (c) The validation loss changes during the training process. (d) The precision score, recall score, MAP at 50-95 score, and F1 score during the training process. (e) The prediction result is shown as a confusion matrix on the test dataset. (f) The test results of the detection model include bounding boxes, classes with confidence scores, and instance segmentation. In (i)-(vi), the figures show the single-component detection task. The multiple components detection result is shown in (vii).}
	\label{Phone sample}
\end{figure*}

Normal cases (plastic), screens (glass), and films (plastic) are recognized as low-value components. They are directed to the low-value recycle process. The middle layer and iPhone case are made of metal and contain electronic components inside, which are considered high-value components. Additionally, batteries are always attached to these two high-value components with different structures and directions. The iPhone case goes through the flipping process first, and then it is sent along with the middle layer to the battery removal system. After the extraction process, they are sorted into the high-value bin without batteries. The segmentation masks clearly show the boundaries between different layers and indicate the direction of each component. The direction information is used to guide the flipping and placing process. The sample of the phone component layers is shown in Figure~\ref{Phone sample} (a).

The dataset is split into 85\% for training and 15\% for validation. The total amount of segmentation annotations in this dataset is 6939. To enhance the robustness of the training dataset, we have applied random cropping, rotation, and flipping as data augmentation techniques to the training dataset. A total of 5,367 images are used to train our model. For validation, at least two unique unseen images from each phone's layers are included to evaluate the model's generalization capability. Furthermore, an additional 300 images are prepared as the test dataset to comprehensively evaluate the model's performance.

\subsubsection{Detection Model Training and Evaluation}
We have implemented transfer learning to use the pre-trained weights for the YOLOv8 segmentation model from the Ultralytics repository. We run the training and testing on a desktop with an Intel 14700K CPU and one NVIDIA GeForce RTX 4070 Super GPU. During the training, we train 100 epochs with a batch size of 16. We have measured the loss of three objectives in our model: the bounding box loss, the segmentation loss, and the classification loss. The loss during the training process can be found in Figure~\ref{Phone sample} (b) and (c).

We use precision, recall, F1 score, and mean average precision (mAP) at $IoU = 0.5-0.95$ (mAP50-95) to quantify the model's performance. Precision quantifies the proportion of true positives among all predicted positives. Recall calculates the proportion of the true positives among all actual positive factors, which is referred to as the true positive rate. The F1 score is the harmonic mean of precision and recall values, providing a balanced assessment of a model's performance. Similarly, mAP represents the average area under the precision-recall curve across multiple classes, which reflects the overall performance of the model. Among that, the intersection of union (IoU) plays a vital role in measuring the overlap between the prediction mask and the ground truth mask in the segmentation model. For example, the threshold at $IoU = 0.5$ means the prediction is classified as a true positive when the IoU score is equal to or larger than $0.5$. $mAP_{50}$ and $mAP_{95}$ are the mean average precision at the IoU thresholds of 0.5 and 0.95. As shown in Figure~\ref{Phone sample} (d), the precision and recall scores reach 0.988, the mAP @ 50-95 score reaches 0.98, and the F1 score reaches 0.986 in the training process. 

The model's performance on the test dataset is evaluated using a confusion matrix, as shown in Figure~\ref{Phone sample} (e). The overall accuracy of this model at $IoU = 0.7$ reaches $98.9\%$. We also show the sample prediction result in Figure~\ref{Phone sample} (f). The confidence score represents the model's certainty whether a detected object exists and belongs to its correct class. In practice, we set the confidence threshold to 0.8 to minimize detection errors.

Another critical factor is the inference time. The system must perform real-time object recognition and process data without interrupting robotic operations. The same desktop is used to test the inference time with a batch size of 1. The average inference time is 19.7 ms, which can meet the requirement for real-time operation.

\subsection{Battery Extraction System with Cooler}
We use nitrogen to simulate the chiller process for cooling phone parts before the separation process. We validate the method's consistency with different phones. This process takes 30 seconds and leaves little or no residual glue on the battery across all the test phones. Additionally, the system is designed to distribute and process multiple phones simultaneously to increase efficiency. The chilling process does not affect the system's output.

\begin{figure*}[htbp!]
\centering
\begin{minipage}[t]{1\textwidth}
    \renewcommand{\arraystretch}{1.1}
    \small
    \captionof{table}{Techno-economic assessment analysis for the proposed system.}
    \label{tab:economic_results}
    \vspace{-4pt}
    \noindent\rule{0.98\textwidth}{0.8pt}
    \begin{tcolorbox}[
      enhanced,
      width=0.98\textwidth,
      colback=gray!25,         
      colframe=gray!25,        
      boxrule=0.2pt,           
      sharp corners,           
      boxsep=0pt,              
      left=0mm, right=2mm, top=1.5mm, bottom=1.5mm, 
      before skip=0mm, 
      after skip=0mm]
        \textbf{Sub-table 1: CAPEX and OPEX cost of the system.}
        \end{tcolorbox}
        \begin{tabularx}{0.98\textwidth}{@{}>{\raggedright\arraybackslash}p{3cm} *{4}{C}@{}}
        \toprule
        \textbf{Category} & \textbf{Capital Costs (USD)} & \textbf{Power Usage (W)} & \textbf{Energy Costs (USD/year)} & \textbf{Maintenance Cost (USD/year)} \\
        \midrule
        Structural Equipment & 10000 & 500 & 205.56 & 200 \\
        Compressor & 820 & 1050 & 431.67 & 16.4 \\
        Stepper Motors & 1000 & 1050 & 431.67 & 20 \\
        Chiller & 25000 & 200 & 82.22 & 500 \\
        Cutting System & 5500 & 2400 & 986.68 & 110 \\
        Working Area & 1000 & 0 & 0 & 20 \\
        Battery Remover & 2000 & 0 & 0 & 40 \\
        Gripper & 3700 & 0 & 0 & 74 \\
        UR5e & 31062 & 200 & 82.22 & 621.24 \\
        Computer & 2996 & 100 & 41.11 & 59.92 \\
        Camera & 334 & 50 & 20.55 & 6.68 \\
        \midrule
        \textbf{Total} & \textbf{83412} & \textbf{5550} & \textbf{2281.71} & \textbf{1668.24} \\
        \bottomrule
    \end{tabularx}
    \begin{tcolorbox}[
      enhanced,
      width=0.98\textwidth,
      colback=gray!25,         
      colframe=gray!25,        
      boxrule=0.2pt,           
      sharp corners,           
      boxsep=0pt,              
      left=0mm, right=2mm, top=1.5mm, bottom=1.5mm, 
      before skip=0mm,     
      after skip=0mm]
        \textbf{Sub-table 2: Economic results comparison (Annual).}
        \end{tcolorbox}
    \begin{tabularx}{0.98\textwidth}{@{}>{\raggedright\arraybackslash}p{3cm} *{4}{C}@{}}
    \toprule
    \textbf{Category} & \textbf{Hourly Throughput (PCs)} & \textbf{Yearly Throughput (lbs)} & \textbf{Total Revenue (USD)} & \textbf{Total Profit (USD)}\\
    \midrule
    \textbf{Proposed System} & \textbf{120} & \textbf{106,546.55} & \textbf{147,034.24} & \textbf{55404.21}\\
    Traditional Process & 6.16 & 5755 & 7941.9 & (76058.1)\\
    \bottomrule
    \end{tabularx}
    \begin{tcolorbox}[
      enhanced,
      width=0.98\textwidth,
      colback=gray!25,         
      colframe=gray!25,        
      boxrule=0.2pt,           
      sharp corners,           
      boxsep=0pt,              
      left=0mm, right=2mm, top=1.5mm, bottom=1.5mm, 
      before skip=0mm, 
      after skip=0mm]
        \textbf{Sub-table 3: Economic results comparison (USD per lb input).}
        \end{tcolorbox}
    \begin{tabularx}{0.98\textwidth}{@{}>{\raggedright\arraybackslash}p{3cm} *{4}{C}@{}}
    \toprule 
    \textbf{Category} & \textbf{Cost (USD/lb input)} & \textbf{Positive Revenue (USD/lb input)} & \textbf{Profit with Supervision (USD/lb input)} & \textbf{Profit w/o Supervision (USD/lb input)} \\
    \midrule
    \textbf{Proposed System} & \textbf{(0.86)} & \textbf{1.38} & \textbf{0.52} & \textbf{1.35}\\
    Traditional Process & (14.60) & 1.38 & (13.22) & N/A \\
    \bottomrule
    \end{tabularx}
\end{minipage}
\end{figure*}

\subsection{Techno-economic Assessment Analysis}
Each component takes approximately 7 seconds to sort, keeping the total processing time per phone under 30 seconds. A comparative techno-economic assessment (TEA) is conducted to evaluate the efficiency and cost of our system relative to the baseline manual process at Sunnking. It considers workload, labor wages, and system deployment costs.

The total cost of our proposed system comprises capital expenditures (CAPEX) and operational expenditures (OPEX). CAPEX represents the cost of hardware components and associated installation. The annual CAPEX is calculated over a 20-year service period. The OPEX of our system includes energy, maintenance, and labor costs per year. The detailed costs are \$2,281.72, \$1,668.24, and \$84,000, respectively. Labor costs are estimated based on an 8-hour workday, 300 workdays per year, and the standard stipend rates at Sunnking. The manual process only includes labor costs. Labor costs are assumed to be the same across our system and manual processes. The proposed system only operates with the operator's working schedule. 
Revenue per phone is estimated using proprietary component value data from the recycling partner. The revenue includes the value of multiple components: battery, board, frame, and screen. We assume both the proposed system and the manual process achieve the same disassembly quality and form, ensuring equal revenue across the two methods.
Results indicate that automation is significantly more economical than manual disassembly due to its high output volume.
The traditional manual process results in a net loss of $\$(13.22)$ USD/lb of input, while our proposed system results in a profit of $\$0.52$ USD/lb of input. The annual CAPEX and OPEX costs are considered during the profit estimation. The detailed TEA analysis for the proposed system is shown in Table~\ref{tab:economic_results}.
Labor costs are the primary reason for the net loss of Sunnking's manual disassembly process. 
The average manual disassembly time takes approximately 9.6 minutes per phone, combined with a labor rate of $\$35$ USD/hour. Given the same labor costs, our proposed automated system significantly escalates the processing speed from 9.6 minutes to 30 seconds per phone. The efficiency improvement makes the entire process profitable. Labor is also a critical reason that the proposed system focuses on the functional layer level disassembly. Any extended and detailed disassembly for small components above the functional layer level goes beyond the capability of current robotic systems. It could introduce more labor costs that exceed the marginal revenue from these components, which could reduce the net profit of the whole process.
Nevertheless, we still need one human to supervise our automated system at this stage. This indicates the potential to further automate the disassembly process and improve economics. Assuming no labor is required, the profit can be increased to $\$1.35$ USD/lb of input. Additionally, the system's throughput capacity could be enhanced by a factor of three.

\section{Conclusions and Discussions}
A fully automated selective disassembly and sorting system for EoL phones has been introduced in this paper, integrating three core processes: (i) an adaptive cutting mechanism that effectively separates phone layers, (ii) a vision-based robotic sorting module employing deep learning for precise component detection and localization, and (iii) a battery removal system utilizing chiller combined with a force mechanism for safe battery extraction. Experimental validation demonstrated a detection accuracy of 98.9\% and a processing throughput exceeding 120 units per hour, thereby efficiently isolating high-value components for downstream recycling. Compared to the current workflow, the automated system brings the disassembly revenue from $\$(13.22)$ net loss to $\$0.52$ profit per pound of EoL phones as input. This estimation considers a human supervisor working on placing the collection unit into the system and monitoring the performance of the system, which can be further replaced with another robotic system to improve efficiency and overall working time. 

This system focuses on providing a robust automated solution across diverse phone types and brands, aiming to meet the requirements of current industrial practice. Current phones, designed and produced over the past 5 to 10 years, have not incorporated eco-design principles for easy disassembly. Regulations such as water resistance and weight restrictions have led to compact designs with glued seals, which hinder the efficiency requirements for non-destructive disassembly. With the implementation of regulations such as Ecodesign for Sustainable Products and advancements in robotics, the cutting system could eventually evolve into a robotic, non-destructive disassembly system.

\bibliographystyle{ieeetr}
\bibliography{Reference}

\end{document}